\documentclass[letterpaper, 10 pt, conference]{ieeeconf}  
\pdfoutput=1

\IEEEoverridecommandlockouts                              

\usepackage{amsmath}
\usepackage{array}
\usepackage{url} 
\usepackage{enumerate}
\usepackage{longtable}
\usepackage{multirow} 
\usepackage{rotating} 
\usepackage{float}
\usepackage{paralist}
\usepackage{amsfonts}
\usepackage{caption}
\usepackage{longtable}

\newcolumntype{L}[1]{>{\raggedright\let\newline\\\arraybackslash\hspace{0pt}}m{#1}}
\newcolumntype{C}[1]{>{\centering\let\newline\\\arraybackslash\hspace{0pt}}m{#1}}
\newcolumntype{R}[1]{>{\raggedleft\let\newline\\\arraybackslash\hspace{0pt}}m{#1}}

\usepackage{graphicx}

\usepackage{times} 

\usepackage{amsmath} 
\usepackage{amssymb}  
\usepackage{amsfonts}
\usepackage{nicefrac}
\usepackage{caption}
\usepackage{subcaption}
\usepackage{algorithm} 
\usepackage{algpseudocode}
\usepackage{wrapfig}

\usepackage{url}
\usepackage[compress]{cite}
\usepackage{paralist}

\graphicspath{{images/}}
\usepackage{ifpdf}
\ifpdf\DeclareGraphicsExtensions{.pdf,.png,.jpg}\fi

\usepackage{todonotes}

\newcommand{\bma}{\begin{bmatrix}}
\newcommand{\ema}{\end{bmatrix}}

\long\def\comment#1{}

\newcommand{\nonote}[1]{}

\usepackage{color}
\definecolor{Junaed_color}{RGB}{0,0,0} 
 
\newcommand{\baad}[1]{} 

\title{\LARGE \bf Dynamic Reconfiguration of Mission Parameters in Underwater Human-Robot Collaboration}

\author{Md Jahidul Islam$^{1}$, Marc Ho$^{2}$, and Junaed Sattar$^{3}$
\thanks{				
 The authors are with the Department of Computer Science and Engineering, University of Minnesota-Twin  Cities, MN 55455, USA. \newline
 {\tt\small E-mail:$\{ ^{1}$islam034, $^{2}$hoxxx323, $^{3}$junaed$\}$@umn.edu} 
}
}

\begin{document}

\maketitle
\thispagestyle{empty}
\pagestyle{empty}

\begin{abstract}
This paper presents a real-time programming and parameter reconfiguration method for autonomous underwater robots in human-robot collaborative tasks. Using a set of intuitive and meaningful hand gestures, we develop a syntactically simple framework that is computationally more efficient than a complex, grammar-based approach.  
In the proposed framework, a convolutional neural network is trained to provide accurate hand gesture recognition; subsequently, a finite-state machine-based deterministic model performs efficient gesture-to-instruction mapping and further improves robustness of the interaction scheme. The key aspect of this framework is that it can be easily adopted by divers for communicating simple instructions to underwater robots without using artificial tags such as fiducial markers or requiring memorization of a potentially complex set of language rules. 
Extensive 
experiments are performed both on 
field-trial data and through simulation, which demonstrate the robustness, efficiency, and portability of this framework in a number of different scenarios. 
Finally, a user interaction study is presented that illustrates the gain in the ease of use of our proposed interaction framework compared to the existing methods for the underwater domain.
%
%
\end{abstract}
\section{Introduction} 
\label{sec:intro}
\begin{figure}[ht]
\centering
\includegraphics [width=\linewidth]{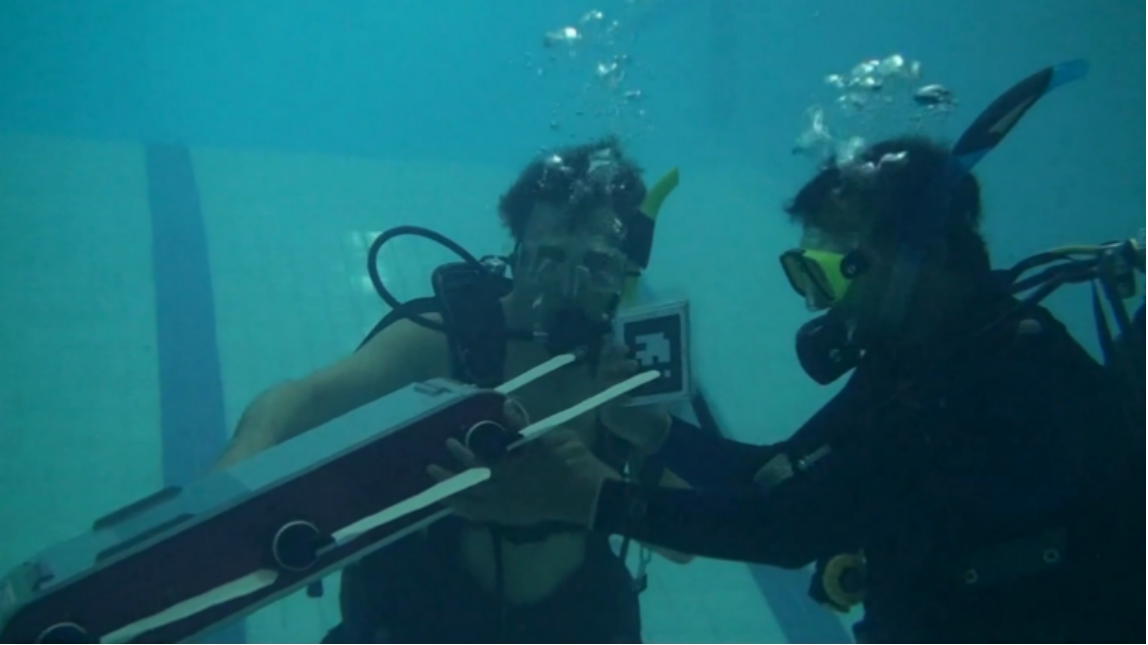}
\vspace{-4mm}
\caption{Divers programming an AUV using the RoboChat~\cite{dudek2007visual} language using ARTag~\cite{fiala2005artag} markers; note the thick ``tag book'' being carried by the diver, which, while necessary, adds to the diver's cognitive load and impacts mission performance}.
\vspace{-4mm}
\label{tag_programming}
\end{figure}

Underwater robotics is an area of significantly increasing importance and applications, and is experiencing a rapid rise in research endeavors. Truly autonomous underwater navigation is still an open problem, with the underwater domain posing unique challenges to robotic sensing, perception, navigation, and manipulation. However, a simple yet robust human-robot communication framework \cite{dudek2007visual, xu2008natural, chiarella2015gesture} is desired in many tasks which requires the use of autonomous underwater vehicles (AUVs).  Particularly, the ability to accept direct human guidance and instructions during task execution (see Fig.~\ref{tag_programming}) is of vital importance.  Additionally, such semi-autonomous behavior of a mobile robot with human-in-the-loop guidance reduces operational overhead by eliminating the necessity of teleoperation (and one or more teleoperators). However, simple and intuitive instruction sets and robust instruction-to-action mapping are essential for successful use of AUVs in a number of critical applications such as search-and-rescue, surveillance, underwater infrastructure inspection, and marine ecosystem monitoring.

The ability to alter parts of instructions (\emph{i.e.}, modifying subtasks in a larger instruction set) and reconfigure program parameters is often important for underwater exploration and data collection processes. Because of the specific challenges in the underwater domain, what would otherwise be straightforward deployments in terrestrial settings often become extremely complex undertakings for underwater robots, which require close human supervision. Since Wi-Fi or radio (\emph{i.e.}, electromagnetic) communication is not available or severely degraded underwater \cite{dudek2008sensor}, such methods cannot be used to instruct an AUV to dynamically reconfigure command parameters. The current task thus needs to be interrupted, and the robot needs to be brought to the surface in order to reconfigure its parameters. This is inconvenient and often expensive in terms of time and physical resources. Therefore, triggering parameter changes based on human input while the robot is underwater, without requiring a trip to the surface, is a simpler and more efficient alternative approach.    

Controlling a robot using speech, direct input (\emph {e.g.}, a keyboard or joystick), or free-form gestures is a general paradigm \cite{coronado2017gesture,chen2015hand,wolf2013gesture} in the context of Human-Robot Interaction (HRI). Unlike relatively less challenging terrestrial environments, the use of keyboard or joystick interfaces or tactile sensors is unappealing in underwater applications since it entails costly waterproofing and introduces an additional point of failure. Additionally, since speech or RGB-D (\emph{i.e.}, visual and depth image)-based interfaces, such as a Leap Motion\texttrademark\ or Kinect\texttrademark\, are not feasible underwater, vision-based communication schemes are more natural for diver-robot interaction.     

This work explores the challenges involved in designing a hand gesture-based human-robot communication framework for underwater robots. In particular, a simple interaction framework is developed where a diver can use a set of \emph{intuitive} and meaningful hand gestures to program the accompanying robot or reconfigure program parameters on the fly. A convolutional neural network-based robust hand gesture recognizer is used with a simple set of gesture-to-instruction mapping. A finite-state machine based interpreter ensures predictable robot behavior by eliminating spurious inputs and incorrect instruction compositions. 

\section{Related Work} \label{sec:related}
Modulating robot control based on human input in the form of speech, hand gestures, or keyboard interfaces has been explored extensively for terrestrial environments \cite{coronado2017gesture,chen2015hand,wolf2013gesture, skubic2004spatial}. However, most of these human-robot communication modules are not readily applicable in underwater applications due to environmental and operational constraints \cite{dudek2008sensor}. Since visual communication is a feasible and operationally simpler method, a number of visual diver-robot interaction frameworks have been developed in the literature. 

A gesture-based framework for underwater visual-servo control was introduced in \cite{dudek2005visually}, where a human operator on the surface was required to interpret the gestures and modulate robot movements. Due to challenging visual conditions underwater \cite{dudek2008sensor} and lack of robust gesture recognition techniques, fiducial markers were used in lieu of free-form hand gestures as they are efficiently and robustly detectable under noisy conditions. In this regard, most commonly used fiducial markers have been those with square, black-and-white patterns providing high contrast, such as ARTags~\cite{fiala2005artag} and April Tags~\cite{olson2011apriltag}, among others. These consist of black symbols on a white background (or the opposite) in different patterns enclosed within a square. Circular markers with similar patterns such as the Photomodeler Coded Targets Module system and Fourier Tags \cite{sattar2007fourier} have also been used in practice.

RoboChat \cite{dudek2007visual} is the first visual language proposed for underwater diver-robot communication. Divers use a set of ARTag markers printed on cards to display predefined sequences of symbolic patterns to the robot, though the system is independent of the exact family of fiducial markers being used. These symbol sequences are mapped to commands using a set of grammar rules defined for the language. 
These grammar rules include both terse imperative action commands as well as complex procedural statements. Despite its utility, RoboChat suffers from two critical weaknesses. Firstly, because a separate marker is required for each \emph{token} (\emph{i.e.}, a language component), a large number of marker cards need to be securely carried during the mission and divers have to search for the cards required to formulate a syntactically correct script; this whole process imposes a rather high cognitive load on the diver. Secondly, the symbol-to-instruction mapping is inherently unintuitive, which makes it inconvenient for rapidly programming a robot. The first limitation is addressed in \cite{xu2008natural}, where a set of discrete motions using a pair of fiducial markers is interpreted as a robot command. Different features such as shape, orientation, and size of these gestures are extracted from the observed motion and mapped to the robot instructions. Since more information is embeddable in each trajectory, a large number of instructions can be supported using only two fiducial markers. However, this method introduces additional computational overhead to track the marker motion and needs robust detection of shape, orientation, and size of the motion trajectory. Furthermore, these problems are exacerbated since both robot and human are suspended in a six-degrees-of-freedom (6DOF) environment. Also, the symbol-to-instruction mapping remains unintuitive.

Since the traditional method for communication between scuba divers is with hand gestures, similarly instructing robots is more intuitive  and flexible than using fiducial markers. Additionally, it relieves the divers of the task of carrying a set of markers, which, if lost, would put the mission in peril. There exists a number of hand gesture-based HRI frameworks \cite{coronado2017gesture, chen2015hand,wolf2013gesture, waldherr2000gesture} for terrestrial robots. In addition, recent visual hand gesture recognition techniques \cite{molchanov2015hand, neverova2014multi,nagi2011max} based on convolutional neural networks have been shown to be highly accurate and robust to noise and visual distortions \cite{fabbri2018enhancing}. A number of such visual recognition and tracking techniques have been successfully used for underwater tracking \cite{shkurti2017underwater} and have proven to be more robust than other purely feature-based methods (\emph{e.g.}, \cite{islam2017mixed}). However, feasibility of these models for hand gesture based diver-robot communication has not been explored in-depth yet. 

\section{Methodology}\label{sec:metho}
The proposed framework is built on a number of components:  the choice of hand gestures to map to command tokens, the robust recognition of hand gestures, and the use of a finite-state machine to enforce command structure and ignore erroneous detections or malformed commands. Each of these components is described in detail in the following sections.

\subsection{Mapping Hand Gestures to Language Tokens}
\label{sec:gesture-to-token}
\begin{figure}[ht]
\centering
\vspace{-3mm}
\includegraphics [width=\linewidth]{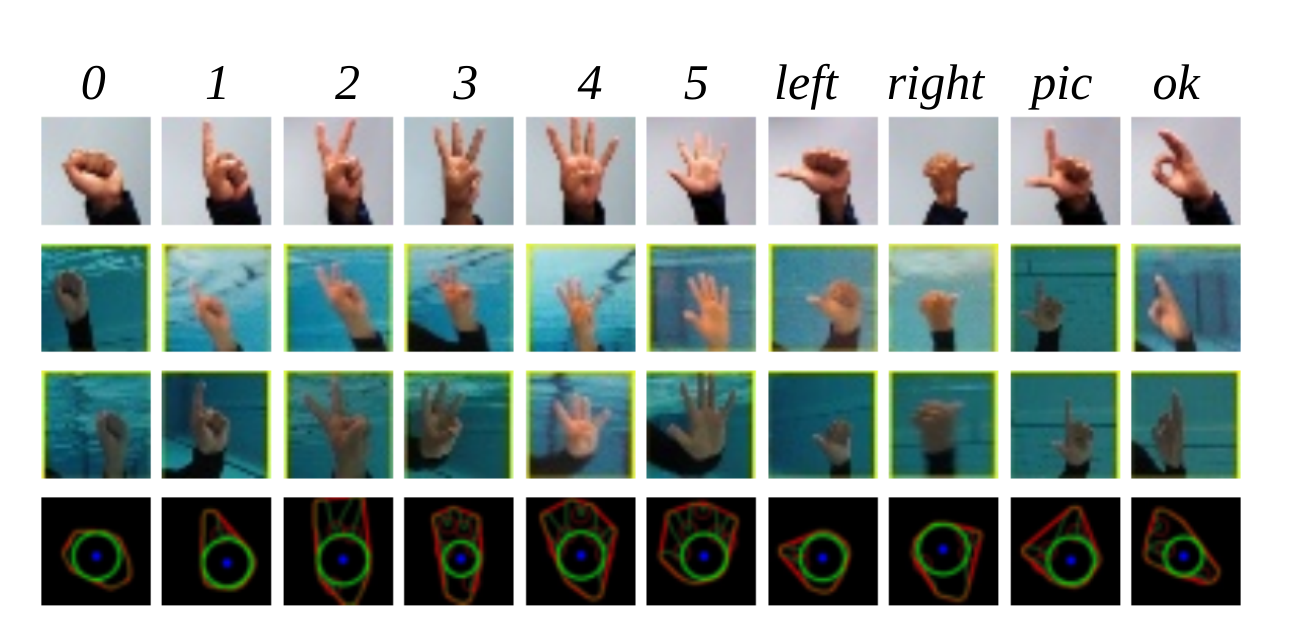}
\vspace{-5mm}
\caption{Hand gestures used in the proposed work. The first three rows show sampled training images for ten different hand gestures in separate columns; the bottom row shows expected hand-contour with different curvature markers for each gesture.}
\vspace{-2mm}
\label{data}
\end{figure}

The key objective of this work is to design a simple, yet expressive framework that can be easily adopted by divers for communicating instructions to the robot without using fiducial markers or memorizing complex language rules. Therefore, we choose a small collection of visually distinctive and intuitive gestures, which would improve the likelihood of robust recognition in degraded visual conditions. Specifically, we use only the ten gestures shown in Fig.~\ref{data}.    

\begin{figure}[ht]
\centering
\vspace{-2mm}
\includegraphics [width=\linewidth]{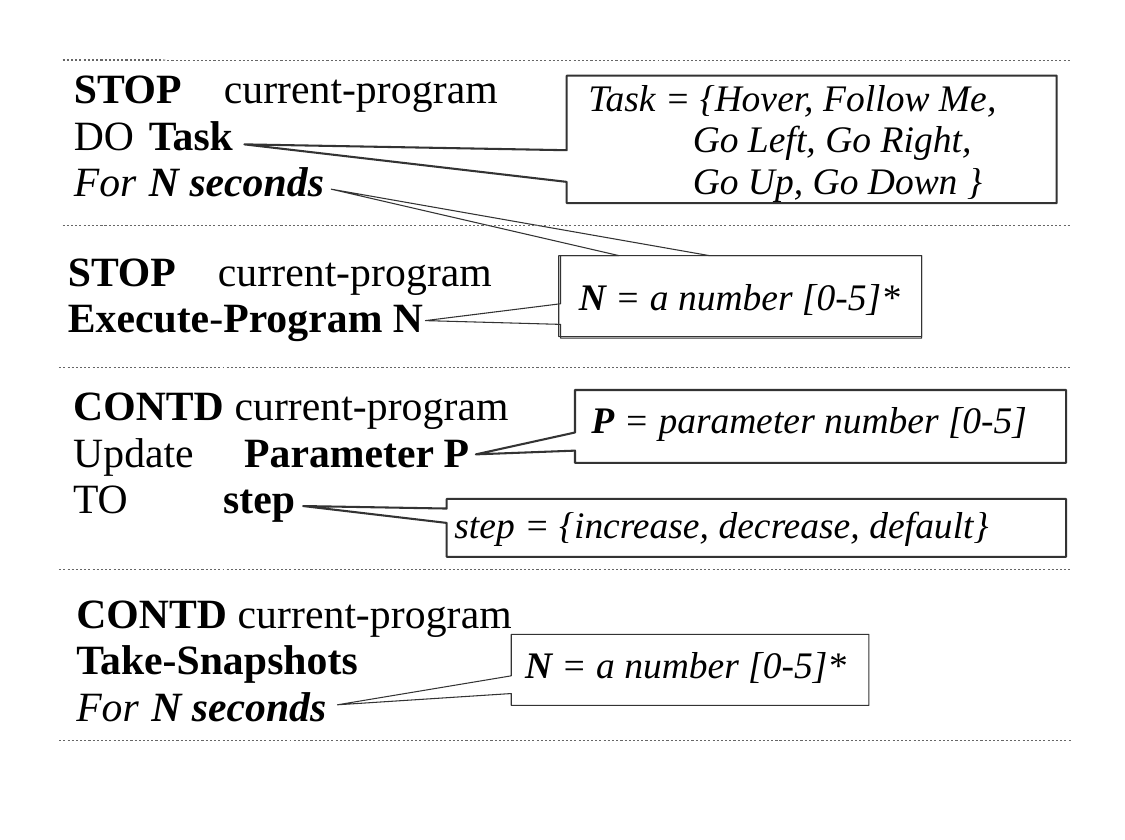}
\vspace{-10mm}
\caption{Set of \textit{task switching} and \textit{parameter reconfiguration} instructions that are currently supported by our framework.}
\label{ins}
\end{figure}

As seen in Fig.~\ref{data}, each gesture is intuitively associated with the command it delivers. Sequences of different combinations of these gestures formed with both hands are mapped to specific instructions. This work concentrates on two different sets of instructions as illustrated in Fig.~\ref{ins}, which are in the following form: 

\begin{itemize}
\item \textbf{Task switching}: This is for instructing the robot to stop executing the current program and start a task specified by the diver, such as hovering, following, or moving left/right/up/down, etc. In other words, these commands are atomic behaviors that the robot is capable of executing. An optional argument can be provided to specify the duration of the new task (in seconds). Another task-switching operation is to stop execution of the current program and start a new program; the difference here is that the robot switches from one sequence of instructions to a different sequence of instructions, rather than just executing an atomic behavior. An operational requirement is that desired programs need to be numbered and known to the robot beforehand.

\item \textbf{Parameter reconfiguration}: This is to instruct the robot to continue the current program with updated parameter values. This enables underwater missions to continue unimpeded (as discussed in Section~\ref{sec:intro}), without interrupting the current task or requiring the robot to be brought to the surface. Here, the requirement is that the tunable parameters need to be numbered and their choice of values need to be specified beforehand. The robot can also be instructed to take pictures (for some time) while executing the current program. This is important for underwater missions, as this facilitates visual logging as the robot executes a preset mission. Also, other sensory data can easily be logged in the same mechanism through a simple extension of the command triggered by a gesture.
\end{itemize}

\begin{figure}[ht]
\centering
\includegraphics [width=\linewidth]{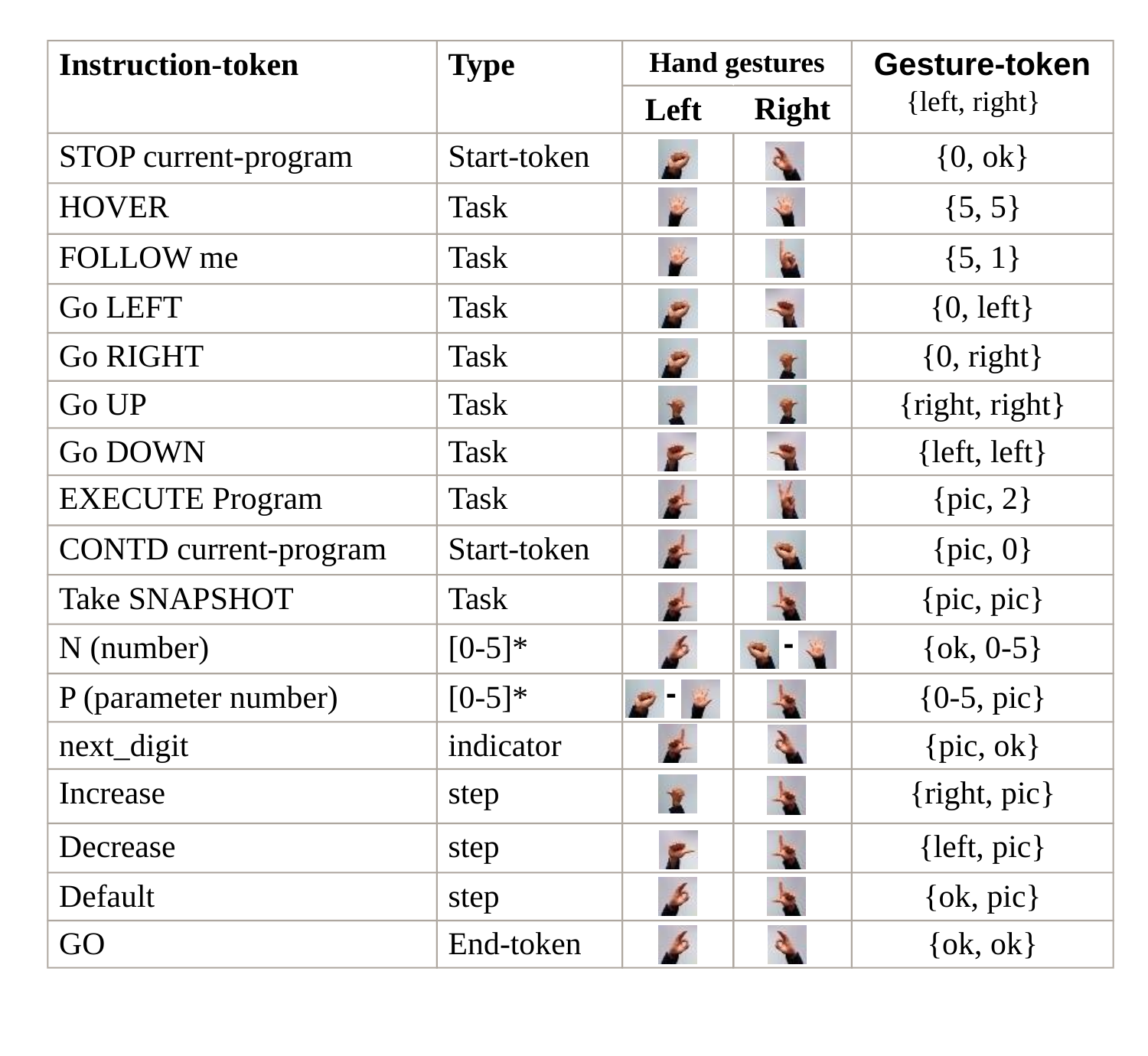}
\vspace{-10mm}
\caption{Mapping of gesture-token to instruction-token used in our framework.}
\label{ins_map}
\vspace{-3mm}
\end{figure}

\begin{figure}[ht]
\centering
\includegraphics [width=\linewidth]{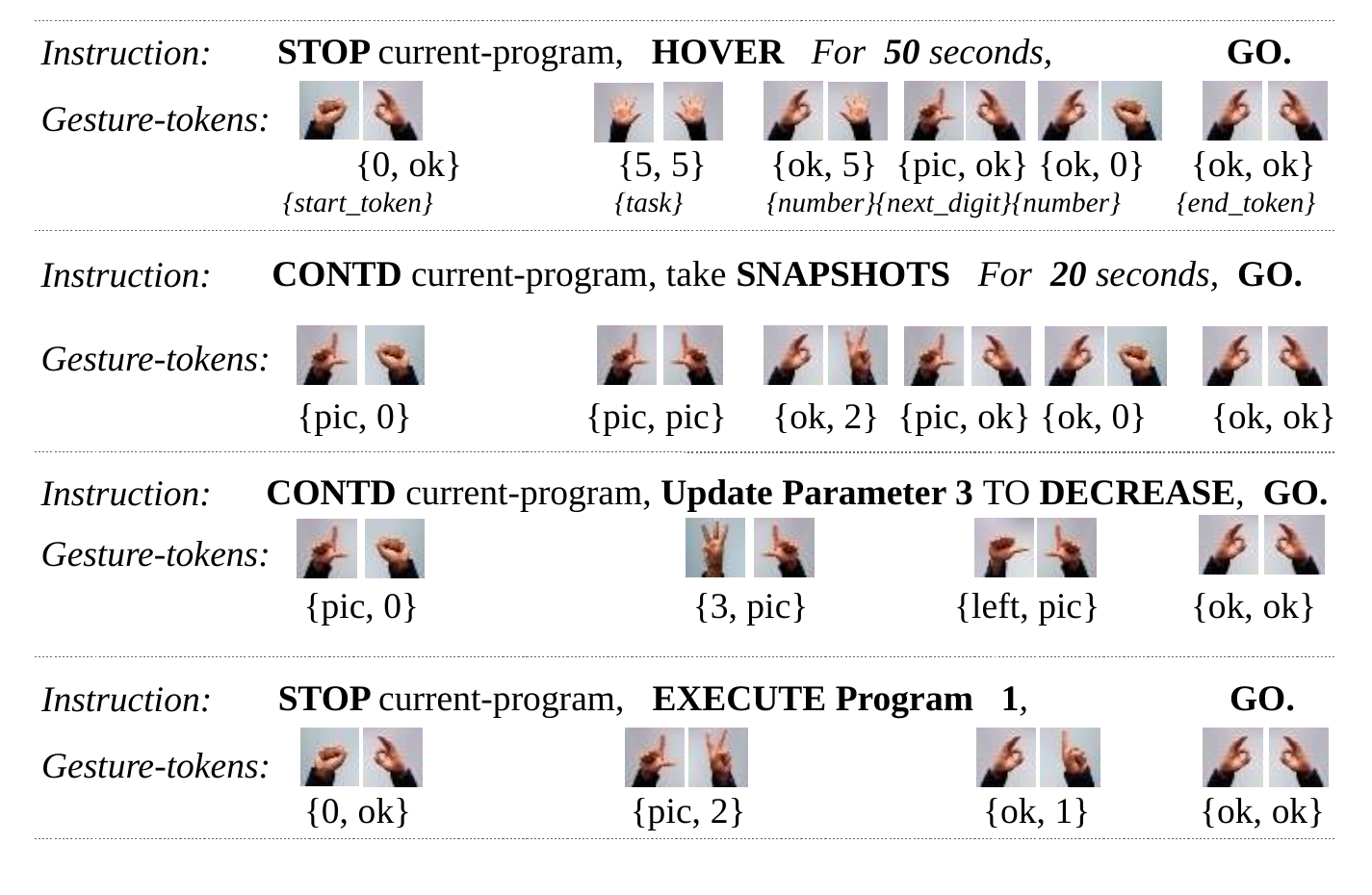}
\vspace{-7mm}
\caption{Some examples of different combinations of hand gestures which are used to generate instructions (based on the gesture-instruction mapping shown in Fig.~\ref{ins_map}).}
\label{exm}
\vspace{-3mm}
\end{figure}

The proposed framework supports a number of task switching and parameter reconfiguration instructions, which can be extended to accommodate more instructions by simply changing or appending a user-editable configuration file. The hand gesture-to-token mapping is carefully designed so that the robot formulates executable instructions only when intended by the diver. This is done by attributing specific hand gestures as \emph{sentinels} (\emph{i.e.}, \textit{start-} or \textit{end-tokens}). Fig.~\ref{ins_map} illustrates the gesture to atomic-instruction mapping used in our framework. Additional examples are shown in Fig.~\ref{exm}, where series of $(start\_token, instruction, end\_token)$ tuples are mapped to corresponding sequences of $gesture\_tokens$. 

\subsection{Hand Gesture Recognition and Instruction Generation}
\begin{figure*}[ht]
\centering
\includegraphics [width=\linewidth]{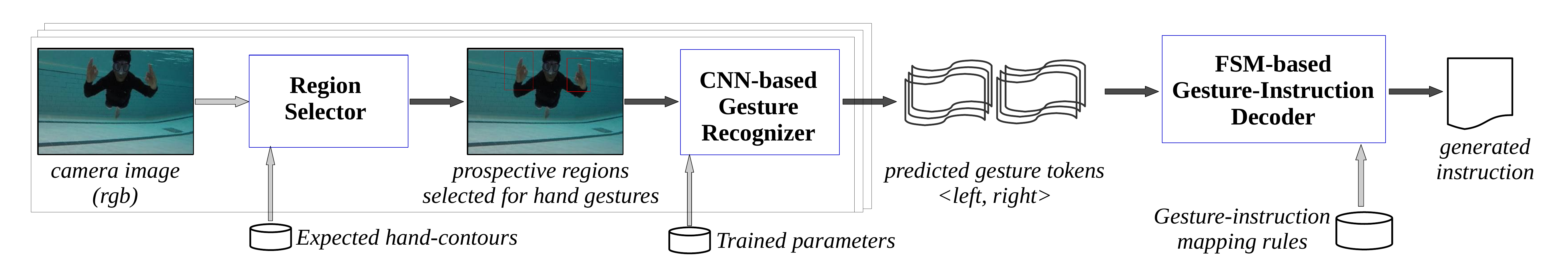}
\vspace{-6mm}
\caption{Overview of the process to map hand gestures to instructions; the left half demonstrates the CNN-based recognition system, whereas the right half depicts the finite-state machine for robust mapping of gestures to instructions.}
\label{Proc}
\vspace{-4mm}
\end{figure*}

Any human-robot communication language must be able to recognize individual tokens robustly, independent of the modality used (\emph{e.g.}, aural, tactile or visual). In the proposed framework, the challenges lie in segmenting the hand gestures from the camera image, accurately recognizing the hand gestures, and then mapping the gestures to instructions. Fig.~\ref{Proc} illustrates the overall process, and the implementation details of each computational component of our framework are described in the following sections.  

\subsubsection{Region Selection} 
\begin{figure}[ht]
\centering
\vspace{-4mm}
\includegraphics [width=\linewidth]{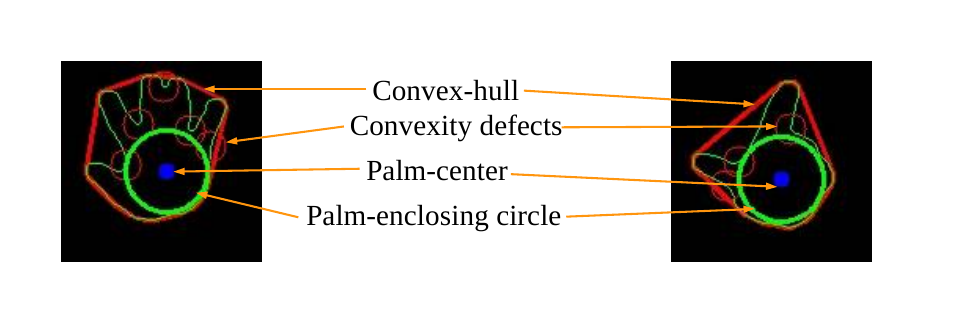}
\vspace{-10mm}
\caption{Two examples of hand contours possessing different contour properties; the left image corresponds to gesture `5' while the right image corresponds to gesture `pic'.}
\label{hcont}
\vspace{-4mm}
\end{figure}

To detect gestures, the hand regions need to be cleanly extracted from the image, which can be challenging in underwater visual conditions which are often degraded. These regions are rectangular but vary in size as divers can be at different distances from the robot. One possible approach is to slide a rectangular window at multiple scales sequentially over the image and attempt hand detections; however, trying each possible rectangular region in such a brute-force fashion is not feasible due to computational and real-time operational constraints. Instead, the following approach is adopted:

\begin{enumerate}
\item First, the camera image (in RGB space) is blurred and thresholded (in HSV-space) for skin-color segmentation \cite{oliveira2009skin}. Here, we assume the diver performs gestures with bare hands; if the diver is to wear gloves, the color thresholding range in the HSV space needs to be adjusted accordingly.

\item Contours of the different segmented regions in the filtered image space are then extracted (see Fig.~\ref{hcont}). Subsequently, different contour properties such as convex hull boundary and center, convexity defects, and important curvature points are extracted. We refer readers to \cite{yeo2015hand} for details about properties and significance of these contour properties.

\item Next, outlier regions are rejected using cached information about the scale and location of hand gestures detected in the previous frame. This step is of course subject to availability of the cached information.

\item Finally, the hand contours of potential regions are matched with a bank of hand contours that are extracted from training data (one for each class of hand gestures as shown in the bottom row of Fig. \ref{data}). Final regions for left and right hand gestures are selected using the proximity values of the closest contour match \cite{yeo2015hand}; \textit{i.e.}, the region that is most likely to contain a hand gesture is selected.     
\end{enumerate}

\begin{figure}[ht]
\centering
\vspace{-2mm}
\includegraphics [width=\linewidth]{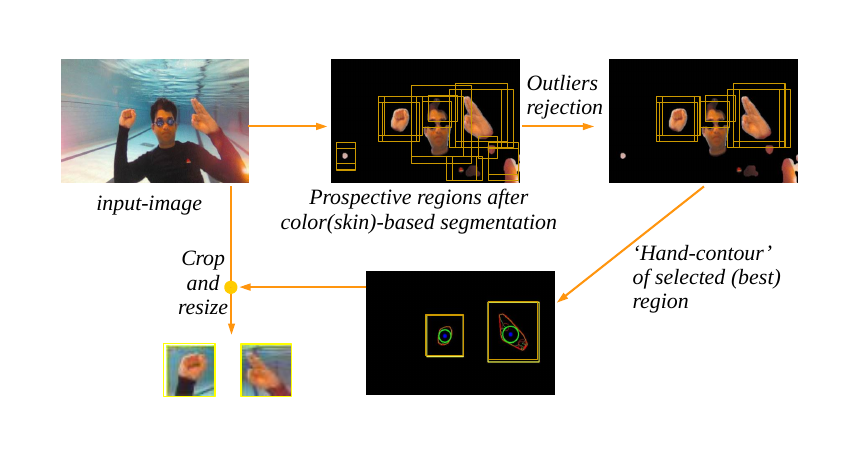}
\vspace{-10mm}
\caption{Outline of the region selection mechanism of our framework: First, the (skin)color-based segmentation is performed to get potential regions for hand gestures; then, outliers are discarded based on cached information about the previous locations of the hands.}
\label{reg}
\vspace{-3mm}
\end{figure} 

\subsubsection{CNN Model for Gesture Recognition} 
\begin{figure*}[ht]
\centering
\includegraphics [width=\linewidth]{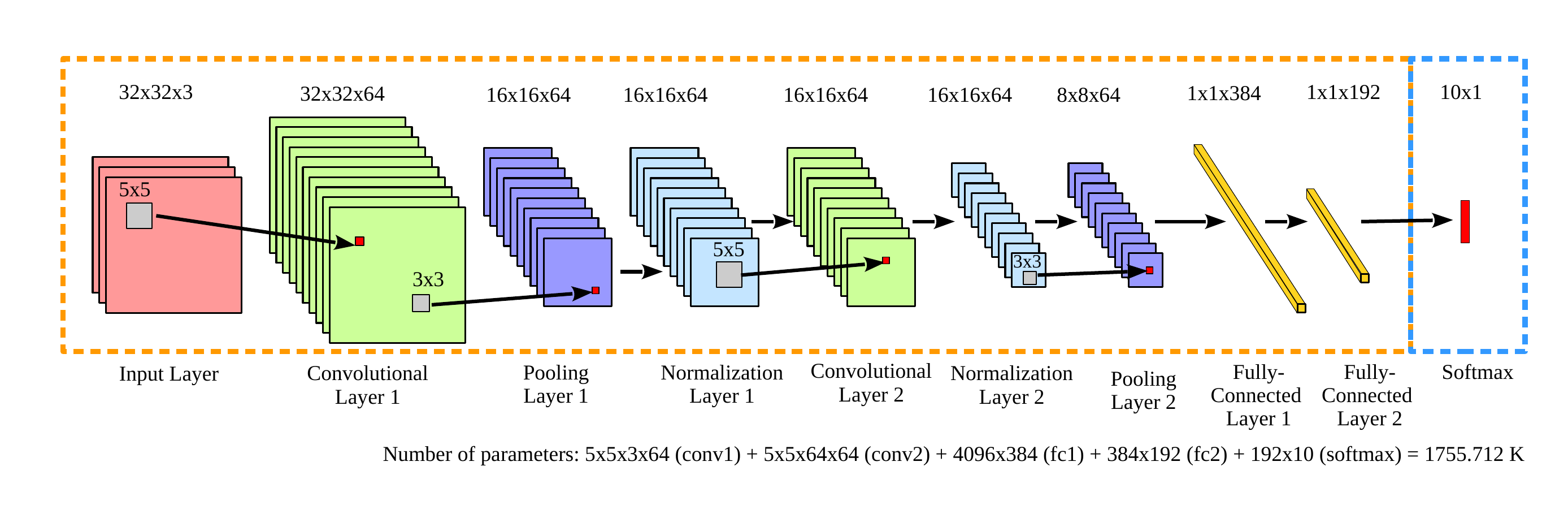}
\vspace{-7mm}
\caption{Architecture of the CNN model used in our framework for hand gesture recognition.}
\label{Arch}
\vspace{-4mm}
\end{figure*}

Following region selection, cropped and resized image-patches of $32\times 32$ are fed to a convolutional neural network (CNN) for gesture recognition. The detailed architecture of the model is illustrated in Fig.~\ref{Arch}. Two convolutional layers are used for extracting and learning the spatial information within the images. Spatial down-sampling is done by max-pooling, while the normalization layer is used for scaling and reentering the data before feeding it to the next layer. The extracted feature vectors are then fed to fully connected layers to learn decision hyperplanes within the distribution of training data. Finally, a soft-max layer provides output probabilities for each class, given input data. Note that similar CNN models have been shown to perform well for small-scale (\textit{i.e.}, $10$-class classification) problems which are similar to ours. 

The dimensions of each layer and the number of parameters are specified in Fig. \ref{Arch}; the details about training and data-sets are provided in Section \ref{sec:Perf}. The trained model is used for classifying hand gestures on $32\times 32$ image patches provided by the \textit{region selection} step. The classified gesture-tokens are passed to a Finite-State Machine (FSM)-based gesture-to-instruction decoder, which we discuss next.

\begin{figure*}[ht]
\centering
\includegraphics [width=\linewidth]{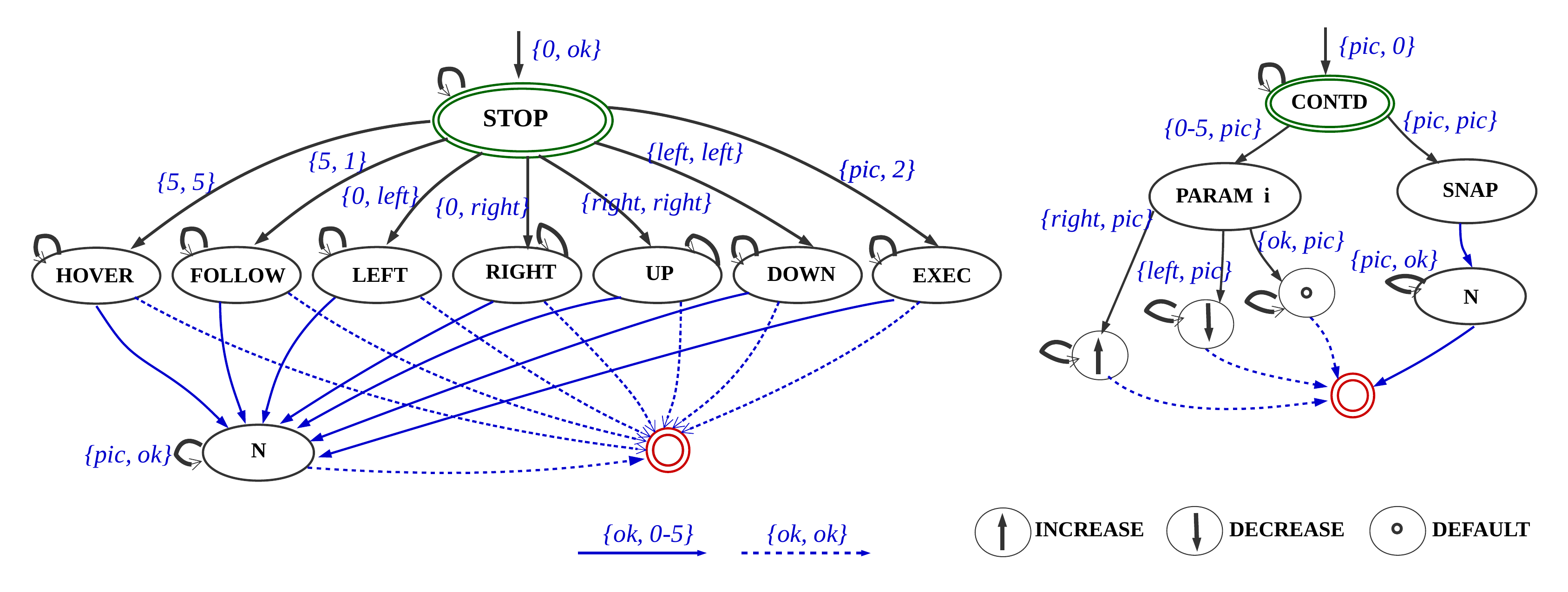}
\vspace{-6mm}
\caption{FSM-based deterministic mapping of gestures-to-instructions (based on the rules defined in Fig. \ref{ins_map}).}
\label{Fsm}
\vspace{-4mm}
\end{figure*} 

\subsubsection{FSM-based Gesture to Instruction Decoder}
An FSM-based deterministic model is used for efficient gesture-to-instruction mapping. As illustrated in Fig.~\ref{Fsm}, the transitions between instruction tokens are defined as functions of gesture tokens based on the rules defined in Fig. \ref{ins_map}. Here, we impose an additional constraint that each gesture token has to be detected for $15$ consecutive frames for the transition to be activated. This constraint adds robustness to missed or wrong classification for a particular gesture token. Additionally, it helps to discard noisy tokens which may be detected when the diver changes from one hand gesture to the next. Furthermore, since the mapping is one-to-one, it is highly unlikely that a wrong instruction will be generated even if the diver mistakenly performs some inaccurate gestures because there are no transition rules other than the correct ones at each state.

\section{Experimental Results}\label{sec:Perf}
We present the experimental results and discuss related implementation details of the proposed framework.

\subsection{Training the CNN Model} 
 The CNN model is implemented using TensorFlow \cite{abadi2016tensorflow} and trained on a Linux machine (CPU) over $35$K RGB-images of hand gestures ($3.5$K for each class). Training data contain $32\times 32\times 3$ images from both underwater and terrestrial environments. A few samples from the training images are shown in Fig. \ref{data} and details about our CNN model are provided in Fig. \ref{Arch}. An additional $4$K images are used for validation and a separate $1$K images are used as a test-set.         

\begin{figure*}[ht]
\centering
\includegraphics [width=\linewidth]{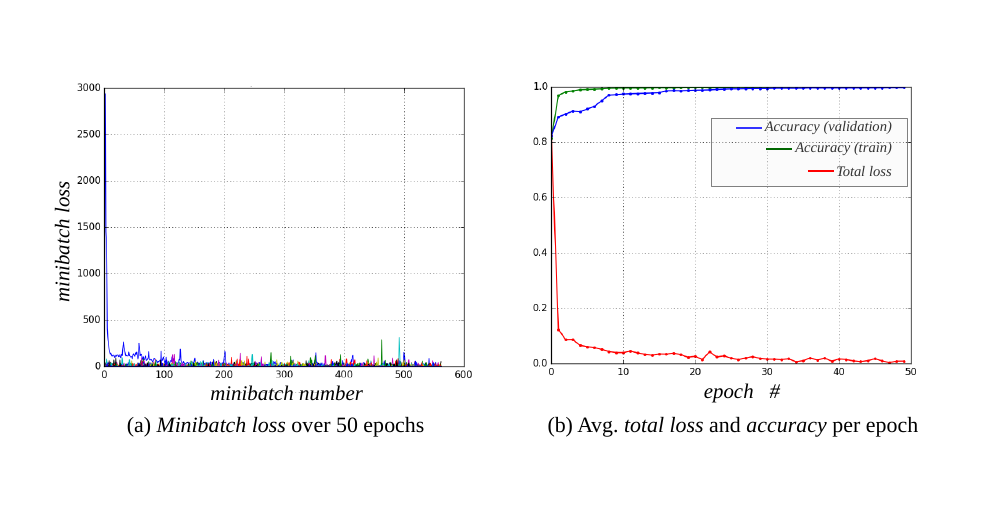}
\vspace{-16mm}
\caption{Training error and accuracy of our CNN model illustrated in Fig. \ref{Arch}; the training and validation sets contain $35$K and $4$K images, respectively; batch-size was set to $128$ while the network was trained for $50$ epochs.}
\label{train}
\vspace{-4mm}
\end{figure*} 

\begin{figure}[ht]
\centering
\vspace{-4mm}
\includegraphics [width=0.9\linewidth]{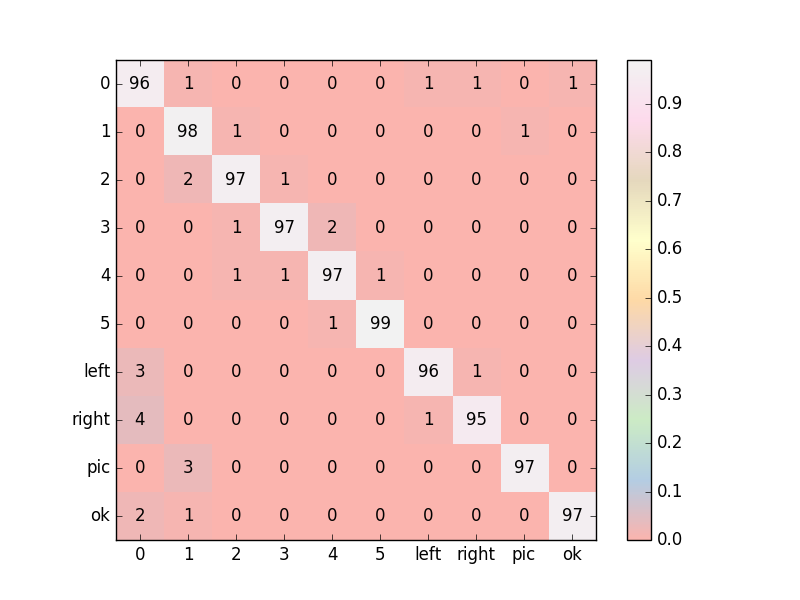}
\vspace{-4mm}
\caption{Confusion matrix based on test-set performance.}
\label{conf}
\vspace{-4mm}
\end{figure}

It takes about two hours to train the model over $50$ epochs after which it reaches a training accuracy of $0.997$. Once the network is trained, model parameters are saved, which are loaded later during testing to perform inference. Fig. \ref{train} illustrates the learning behavior of the network in terms of training loss and classification accuracy. Maximum validation accuracy of $0.986$ is achieved after $50$ epochs of training, and the test-set accuracy is $0.969$. Fig. \ref{conf} shows the confusion matrix based on test-set performance of the model. 

\subsection{Performance Evaluation on Real-World Data}
We used an underwater drone (OpenROV 2.8 \cite{OpenROV}) for our experiments in a closed-water (swimming pool) scenario. Sequences of hand gestures pertaining to different types of instructions are performed by three participants. In addition, synthetic test data is generated by augmenting different combinations of the recorded gesture sequences.    

\begin{figure}[ht]
\centering
\vspace{4mm}
\includegraphics [width=\linewidth]{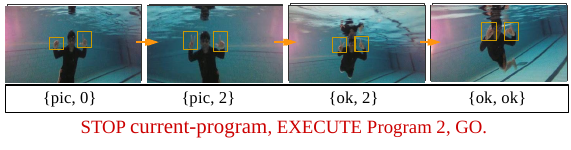}
\vspace{-6mm}
\caption{Generation of instructions from a sequence of hand gestures using our model}
\label{uw}
\vspace{-4mm}
\end{figure}

The test data is used to evaluate the performance of our framework, as demonstrated in Fig. \ref{uw}. There are a total of $30$ sets of image sequences in the test set (each image is $640\times480\times3$). An additional $20$ sets of test data are collected on land to inspect the performance in noise-free visual conditions. Table \ref{comp} summarizes the performance of our framework for both classes of test data. We find that the overall accuracy of the framework mostly depends on region selection; that is, once the hand gestures are correctly segmented out, gesture recognition and gesture-to-instruction mapping demonstrate a high-degree of accuracy. As demonstrated by the bottom row of Table \ref{comp}, our framework successfully decoded all instructions from the noise-free terrestrial data even though gesture recognition accuracy was not perfect (\textit{i.e.}, $0.945$). This is due to the robust FSM-based gesture-to-instruction mapping that ensures the following transition rules: 
\begin{itemize}
\item State transitions are activated only if the corresponding gesture tokens are detected for $15$ consecutive frames.
\item There are no transition rules (to other states) for incorrect gesture tokens. 
\end{itemize}

Consequently, an incorrect recognition has to happen $15$ consecutive frames to generate an incorrect instruction, which is highly unlikely. However, in challenging visual conditions, region selection often fails to segment out the hand gestures correctly, which causes the overall process to fail. As the first row of Table \ref{comp} suggests, our framework fails in $6$ test cases out of $30$. We inspected the failed cases and found the following issues: 

\begin{itemize}
\item Surface reflection and air bubbles often cause problems for the region selector. Although surface reflection is not common in deep water, suspended particles and limited visibility will be additional challenges in deep open-water scenarios.  
\item In some cases, the diver's hand(s) appeared in front of his face or only partially appeared in the field-of-view. In these cases, not all of the hand(s) appeared in the selected region which eventually caused the gesture recognizer to detect `$1$'s as `$0$'s, or `$pic$'s as `$1$'s, etc.    
\end{itemize}

\begin{table}[ht]
\caption{Performance evaluation of our framework based on real-world data.}
\centering
\begin{tabular}{|c|c||c|c|} \hline
Operating & Total \# of & Successfully & Accuracy  \\  
Medium & Instructions (Gestures) & Decoded & (\%)  \\ \hline \hline
Underwater & $30$ ($162$) & $24$ ($128$) & $80$ ($78$)  \\ \hline
Terrestrial & $20$ ($132$) & $20$ ($121$) & $100$ ($94.5$) \\ \hline
\end{tabular}
\label{comp}
\end{table}

\subsection{Performance Evaluation through Gazebo Simulation}
\begin{figure}[ht]
\centering
\vspace{-2mm}
\includegraphics [width=\linewidth]{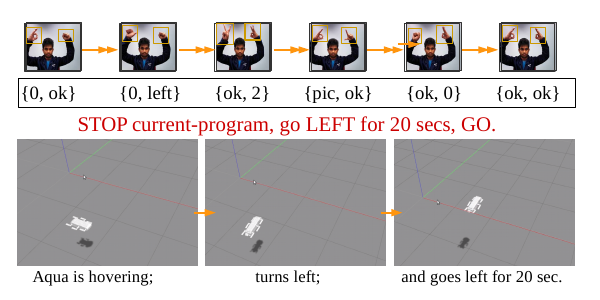}
\vspace{-5mm}
\caption{Controlling an Aqua robot based on instructions generated from sequence of hand gestures performed by a person; the simulation is performed in Gazebo, on ROS-kinetic platform }
\label{simu}
\vspace{-4mm}
\end{figure}

We also performed simulation experiments on controlling an Aqua robot~\cite{dudek2007aqua} based on instructions generated from sequence of hand gestures performed by a participant. The gesture sequences are captured through a webcam and the simulation is performed in Gazebo on the ROS Kinetic platform. As illustrated in Figure \ref{simu}, gesture tokens are successfully decoded to control the robot. Although a noise-free simulation environment does not pose most challenges that are common in the real world, it does help set benchmarks for expected performance bounds and is useful in human interaction studies, which is described in the following section.

\subsection{Human Interaction Study}
We performed a human interaction study where the participants are introduced to our hand gesture based framework, the fiducial-based RoboChat framework~\cite{dudek2007visual}, and the RoboChat-Gesture framework \cite{xu2008natural} where a set of discrete motions from a pair of fiducials are interpreted as gesture-tokens. AprilTags~\cite{olson2011apriltag} were used for the RoboChat trials to deliver commands.

A total of ten individuals participated in the study, who were grouped according to their familiarity to robot programming paradigms in the following manner: 
\vspace{1mm}
\begin{itemize}
\item Beginner: participants who are unfamiliar with gesture/fiducial based robot programming ($2$ participants)
\item Medium: participants who are familiar with gesture/fiducial based robot programming ($7$ participants)
\item Expert: participants who are familiar and practicing these frameworks for some time ($1$ participant)
\end{itemize}

This approach is similar to the one used by~\cite{xu2008natural}. In the first set of trials, participants are asked to perform sequences of gestures to generate the following instructions (Fig. \ref{exm}) in all three interaction paradigms: 
\vspace{1mm}
\begin{enumerate}[$\hspace{4mm}1.$ ]
\item STOP current-program, HOVER for 50 seconds, GO.
\item CONTD current-program, take SNAPSHOTS for 20 seconds, GO.
\item CONTD current-program, Update Parameter 3 to DECREASE, GO.
\item STOP current-program, EXECUTE Program 1, GO.
\end{enumerate}

The second set of trials, participants had to program the robot with complex instructions and were given the following two scenarios:
\begin{enumerate}[\hspace{4mm}a.]
\item \textit{The robot has to stop its current task and execute program 2 while taking snapshots, and}
\item \textit{The robot has to take pictures for 50 seconds and then start following the user.}
\end{enumerate}

\begin{figure}[ht]
\centering
\vspace{-1mm}
\includegraphics [width=\linewidth]{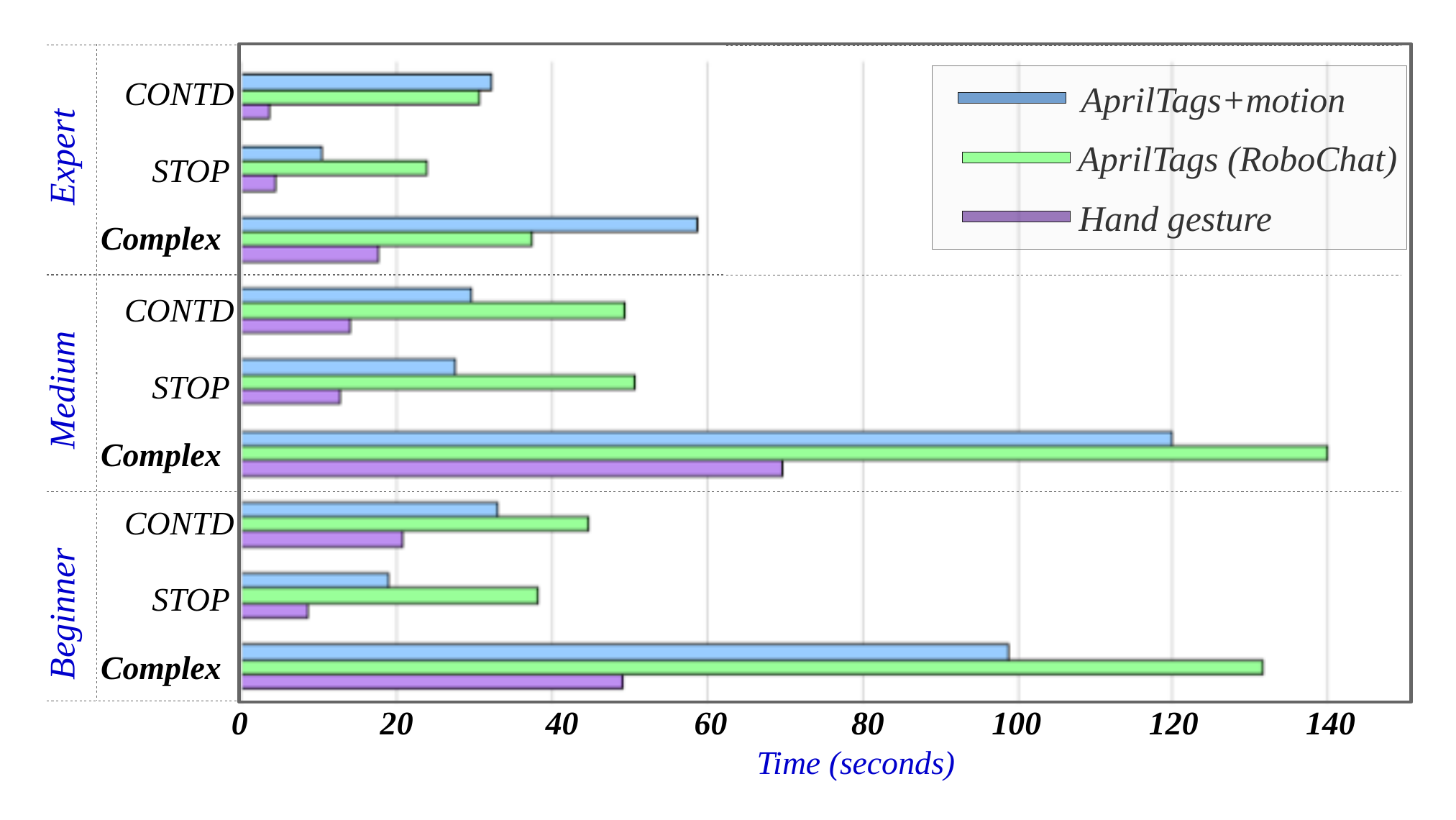}
\vspace{-6mm}
\caption{Comparisons of average time taken to perform gestures for successfully generating different types of programs ($STOP$: instruction $1$ and $4$, $CONTD$: instruction $2$ and $3$, $Complex$: scenario $a$ and $b$) .}
\label{hri}
\vspace{-4mm}
\end{figure} 

For all the experiments mentioned above, participants performed gestures with hands, AprilTags, and discrete motions with AprilTags. Correctness and the amount of time taken were recorded in each case. Fig. \ref{hri} shows the comparisons of average time taken to perform gestures for generating different types of instructions. Participants quickly adopted the hand gestures to instruction mapping and took significantly less time to finish programming compared to the other two alternatives. Specifically, participants found it inconvenient and time consuming to search through all the tags for each instruction token. On the other hand, although performing a set of discrete motions with only two AprilTags saves time, it was less intuitive to the participants. As a result, it still took a long time to formulate the correct gestures for complex instructions, as evident from Fig. \ref{hri}.

One interesting result is that the \textit{beginner} users took less time to complete the instructions compared to \textit{medium} users. This is probably due to the fact that unlike the beginner users, medium users were trying to intuitively interpret and learn the syntax while performing the gestures. However, as illustrated by Table \ref{hri_dudes}, beginner users made more mistakes on average before completing an instruction successfully. The expert user performed all tasks on the first try, hence only a comparison for beginner and medium users is presented. Since there are no significant differences in the number of mistakes for any types of user, we conclude that simplicity, efficiency, and intuitiveness are the major advantages of our framework over the existing methods.   

\begin{table}[ht]
\caption{Average number of mistakes using [$hand$ $gesture$, $Robochat$, $AprilTtags$ $with$ $motion$] for different users before correctly generating the instruction.}
\centering
\begin{tabular}{|c|c||c|c|} \hline
Instruction & Total \#\ of & Beginner & Medium  \\  
Type & Instructions (Gestures) & User & User  \\ \hline \hline
STOP & $2$ $(10$) & $[2,1,3]$ & $[1,0,1]$ \\ \hline
CONTD & $2$ ($10$) & $[0,0,1]$ & $[0,0,0]$   \\ \hline
Complex & $2$ ($16$) & $[2,3,7]$ & $[2,2,3]$   \\ \hline
\end{tabular}
\label{hri_dudes}
\end{table}

\section{Conclusions and Future Work}\label{sec:con}
We present a hand gesture-based human-robot communication framework for underwater robots, where divers can use a set of intuitive and meaningful hand gestures to program new instructions or reconfigure existing program parameters for an accompanying robot on-the-fly. In the proposed framework, a CNN model provides accurate hand gesture recognition and an FSM-based deterministic model performs efficient gesture-to-instruction mapping. Accuracy and robustness of the framework is evaluated through extensive experiments, while a user interaction study is performed to evaluate the usability of the interface. Future work will investigate methods to accommodate a larger vocabulary of instructions and useful features, such as control-flow tokens, while maintaining simplicity and robustness of the approach. In addition, work will focus on designing an improved and more robust region selector. A complete evaluation of the interaction framework through open-water trials is the immediate next step.

\bibliographystyle{plain}
\bibliography{allbibs}

\end{document}